\documentclass[sigconf]{acmart}

\renewcommand\footnotetextcopyrightpermission[1]{}
\AtBeginDocument{%
  }

\setcopyright{acmlicensed}
\copyrightyear{2018}
\acmYear{2018}
\acmDOI{XXXXXXX.XXXXXXX}
\acmConference[WWW '27]
  {Proceedings of the ACM Web Conference 2027}
  {May 10--14, 2027}
  {Dublin, Ireland}
\acmISBN{978-1-4503-XXXX-X/2018/06}
\usepackage{booktabs}
\usepackage{multirow}
\usepackage[table]{xcolor}
\usepackage{array}
\usepackage{tabularx}
\usepackage{hyperref}
\usepackage[most]{tcolorbox}
\definecolor{promptgreen}{RGB}{0,128,0}
\definecolor{promptbackground}{RGB}{245,251,245}

\definecolor{PersonaMemBG}{RGB}{252,229,229}
\definecolor{PrefEvalBG}{RGB}{228,230,252}
\definecolor{PersonaBenchBG}{RGB}{253,239,221}
\definecolor{OverallBG}{RGB}{228,230,252}

\definecolor{rowA}{HTML}{FFBFBF}
\definecolor{rowB}{HTML}{FFCACA}
\definecolor{rowC}{HTML}{FFD5D5}
\definecolor{rowD}{HTML}{FFE0DA}
\definecolor{rowE}{HTML}{FFEBDD}
\definecolor{rowF}{HTML}{FFF5ED}
\definecolor{rowG}{HTML}{FFFAF5}

\usepackage{graphicx}
\usepackage{subcaption}

\usepackage{booktabs}
\usepackage{multirow}
\usepackage[table]{xcolor}
\usepackage{array}
\begin{document}

\title{WMem: Learning to Remember by Modeling Memory Dynamics with World Models}
\settopmatter{authorsperrow=4}
\settopmatter{authorsperrow=1}
\author{%
Huizu Lin$^{1}$,
Tao Huang$^{2}$,
Peixi Wu$^{1}$,
Chengkai Huang$^{3}$,
Zijie Liu$^{1}$,
Xiangyu Yu$^{1}$,\\
Jitong Song$^{4}$,
Yansong Peng$^{1}$,
Hebei Li$^{1,\dagger}$,
Lina Yao$^{3}$,
Xiaoyan Sun$^{1}$\\[3pt]
$^{1}$University of Science and Technology of China, Hefei, China\\
$^{2}$University of Chinese Academy of Sciences, Beijing, China\\
$^{3}$University of New South Wales, Sydney, Australia\\
$^{4}$The University of Hong Kong, Hong Kong, China\\
}

\renewcommand{\shortauthors}{H. Lin et al.}

\begin{abstract}
Large language model (LLM) agents require persistent memory for extended interactions, yet existing memory-update methods struggle to preserve useful information reliably. Heuristic strategies may discard valuable memories, while learning-based approaches discover update policies through trial-and-error feedback. Even successful policies can mistakenly remove important information during individual updates, with such errors going unnoticed when the discarded memories happen not to be needed for downstream answers. Without assessing memory changes at each update, methods struggle to provide fine-grained feedback for reliable memory evolution. To address these challenges, we introduce \textbf{WMem}, a framework that learns memory-update policies through world modeling and step-level predictive feedback. WMem comprises three complementary components: World Modeling for Memory Dynamics, Step-Level Predictive Feedback, and Memory Damage Check. The Memory World Model shares parameters with the policy and predicts whether information will be preserved, revised, or absent after the next update, using supervision from observed trajectories. Step-Level Predictive Feedback leverages prediction accuracy to guide the policy toward more appropriate update decisions for individual memory units, weighting guideline scores by the likelihood of observed memory dynamics. Memory Damage Check assesses changes against dialogue evidence and penalizes unsupported revisions, unjustified omissions, and loss of contextual conditions. Experiments on PersonaMem, PrefEval, and PersonaBench demonstrate substantial overall improvements over strong baselines across extended interaction histories and noisy contexts.
\end{abstract}



\received{20 February 2007}
\received[revised]{12 March 2009}
\received[accepted]{5 June 2009}

\maketitle

\begingroup
\renewcommand{\thefootnote}{\fnsymbol{footnote}}
\footnotetext[2]{Corresponding author}
\endgroup

\section{Introduction}
\label{sec:introduction}

Large language model (LLM) agents require persistent
memory to sustain interactions beyond their context windows.
Early systems address this need through long-term storage
and retrieval~\citep{zhong2023memorybank} and hierarchical
memory management~\citep{packer2023memgpt}.
Subsequent approaches organize accumulated information through
dynamically connected entries~\citep{xu2025amem}, selective
extraction and consolidation~\citep{chhikara2025mem0}, and
reflection-based refinement for multimodal recommendation
agents~\citep{cong2026seeing}.
As interactions continue, however, agents must also decide
which facts to preserve, which content to revise, and
which information to discard.
Recent research therefore treats memory updating as a
learnable decision process.
Reinforcement learning (RL) supports sequential memory
overwriting~\citep{yu2025memagent}, structured memory
operations~\citep{yan2025memoryr1}, and memory construction
guided by downstream task feedback~\citep{wang2025memalpha}.
Other approaches jointly optimize memory consolidation and
reasoning~\citep{zhou2025mem1}, or learn guidelines and
strategies governing memory evolution~\citep{xu2026memcoe,han2026hips}.
Despite these advances, existing learning-based methods still optimize memory updates largely through trial-and-error feedback. Even when the resulting policy performs well, individual updates may mistakenly discard important facts or overwrite valid information. Such errors can remain undetected when the lost information happens not to be needed for downstream answers. Consequently, successful task outcomes may conceal
changes to information unnecessary for the current answer.
Guideline-based scores assess update quality, but do not
explicitly train agents to anticipate how individual
information units evolve.
Modeling these dynamics provides finer-grained supervision
for learning updates that incorporate new evidence
while preserving relevant information and its essential
contextual conditions.

\begin{figure}
    \centering
    \includegraphics[width=\linewidth]{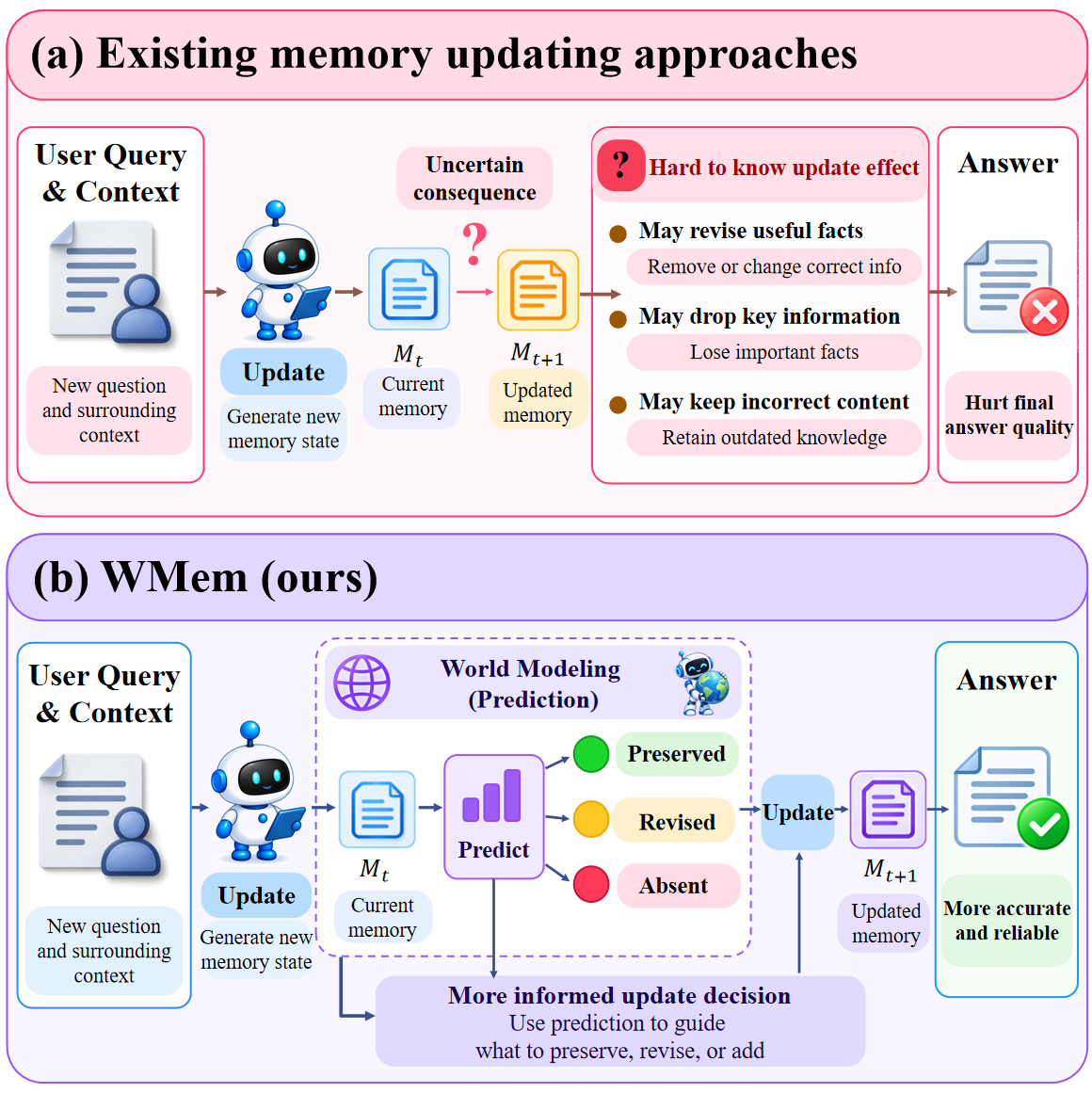}
    \caption{(a) Existing memory updating approaches rely on task-level rewards without explicitly modeling memory evolution, potentially leading to unreliable updates and degraded answer quality. (b) Our WMem predicts memory transitions through world modeling, providing predictive guidance for more informed memory-update decisions and ultimately improving answer accuracy and reliability.}
    \label{Abstract}
\end{figure}

World models offer a framework for establishing this
connection by learning how actions transform states.
Recent advances demonstrate that learned environment dynamics
can support policy optimization through imagined
experience~\citep{hafner2025dreamerv3}.
For language agents, predicting action consequences can
likewise support look-ahead decision-making and policy improvement
through simulated experience~\citep{fang2025webevolver,xiao2026webworld}.
These findings motivate modeling memory evolution
through changes to individual information units.
Each update may preserve facts, revise content,
or remove conditions needed for subsequent reasoning.
Predicting these changes provides finer-grained supervision
beyond update scores and answer rewards.
Through shared parameters, this objective helps
the policy learn to incorporate new evidence
while preserving relevant information and conditions.
To inform policy optimization, such predictions must also
be translated into feedback for update decisions.
Recent work studies action-level feedback~\citep{lin2026auso}, step-level
rewards derived from trajectory preferences~\citep{liu2025istar}, and
advantage assignment through trajectory graphs~\citep{li2026salt}.
Motivated by these approaches, we investigate whether predicted
memory dynamics can provide complementary feedback alongside
guideline-based update scores.
Guideline scores assess update quality, while predictive
likelihoods indicate how well the world model captures
observed information changes.
Combining these complementary signals connects learned
memory dynamics with update-quality assessments,
allowing predictive feedback to inform memory-policy
optimization.
This raises a natural question:
\emph{\textbf{Can anticipating how updates transform stored information
provide useful supervision for learning more effective
agent memory-update policies?}} 

To this end, we introduce \textbf{WMem}, a framework
for learning to remember by modeling memory dynamics
with world models.
As illustrated in Figure~\ref{Abstract}, WMem integrates
three core mechanisms.
(1) \textbf{World Modeling for Memory Dynamics} equips the memory policy with a shared-parameter world
model that predicts whether selected information units
will be preserved, revised, or absent after the next update.
Learning from observed trajectories provides unit-level
supervision, coupling memory-policy learning with
the prediction of how updates reshape information
available for reasoning.
(2) (2) \textbf{Step-Level Predictive Feedback}
uses observed-label likelihoods to adaptively weight
update-level guideline scores, forming a process reward
that complements downstream task feedback.
This connects unit-level dynamics predictions with
update-quality assessments, allowing learned memory
dynamics to inform the trajectory-level optimization
of the memory policy.
(3) \textbf{Memory Damage Check} assesses changes
against the available dialogue evidence, penalizing
unsupported modifications, unjustified omissions,
and loss of essential contextual conditions.
It abstains when evidence is insufficient, complementing
predictive feedback with explicit assessment
of whether memory changes are justified.

Gradient alignment regulates predictive supervision
according to its compatibility with RL optimization,
balancing dynamics learning with policy improvement.
Together, the memory dynamics, the evidence-grounded update
assessment, and downstream task rewards can guide
the learning of effective memory-update policies.
Our contributions are as follows:
\begin{itemize}
\item We propose \textbf{WMem}, a framework that learns
world models of memory dynamics to capture how
information evolves across successive updates,
providing predictive supervision for learning
to preserve and revise stored information.

\item We develop step-level predictive feedback to connect
memory dynamics with policy optimization, complemented
by an evidence-grounded memory damage check that
guards against unsupported revisions, unjustified
information loss, and erosion of essential context.

\item Experiments on PersonaMem, PrefEval, and
PersonaBench demonstrate improvements over strong
baselines, while ablation studies and case analyses
support the effectiveness of the proposed mechanisms.
\end{itemize}

\definecolor{PersonaMemBG}{RGB}{252,229,229}
\definecolor{PrefEvalBG}{RGB}{228,230,252}
\definecolor{PersonaBenchBG}{RGB}{253,239,221}
\definecolor{OverallBG}{RGB}{228,230,252}

\section{Related Work}
\label{sec:related_work}

\subsection{Memory Architectures for LLM Agents}

Memory architectures determine how agents represent,
organize, and access information beyond their context
windows.
MemoryOS coordinates hierarchical storage and
retrieval~\citep{kang2025memoryos}, while MIRIX
distributes memory management across specialized
agents and complementary memory
types~\citep{wang2025mirix}.
Related retrieval architectures include RAPTOR,
which builds trees of recursive summaries to access
information at multiple abstraction
levels~\citep{sarthi2024raptor}.
HippoRAG combines knowledge graphs with Personalized
PageRank~\citep{gutierrez2024hipporag}, and HippoRAG 2
strengthens passage integration and LLM-assisted
retrieval~\citep{gutierrez2025hipporag2}.
A-MEM maintains interconnected notes with evolving
contextual representations~\citep{xu2025amem},
while LightMem separates sensory filtering, short-term
organization, and long-term
consolidation~\citep{fang2025lightmem}.
Nemori further uses prediction error to distill
semantic knowledge from coherent
episodes~\citep{ma2026nemori}.
These approaches emphasize how information is organized,
consolidated, and retrieved.
Our work focuses on learning how memory should change
as new evidence arrives and existing information
must be reconsidered.
We learn a world model of memory dynamics to predict
whether information units will be preserved, revised,
or absent after an update, providing predictive
supervision for memory-policy learning.
Whereas prediction-based knowledge distillation uses
prediction error to extract useful knowledge,
predictive objective models consequences
of memory updates and connects them with policy
learning.

\subsection{Learning Memory Update Strategies}

Learning-based approaches optimize memory management
through task feedback and accumulated experience.
Memory-R1 learns structured memory operations and
answer generation through outcome-driven reinforcement
learning~\citep{yan2025memoryr1}.
MemCoE introduces learned guidelines and guideline-aligned
process rewards~\citep{xu2026memcoe}, while HiPS
co-evolves shared and user-specific strategies
alongside the policy~\citep{han2026hips}.
MemSkill learns to select reusable memory skills
and refines them through difficult
cases~\citep{zhang2026memskill}.
Complementary approaches improve experience reuse:
MemRL learns utility estimates over episodic memories
with a frozen language model~\citep{zhang2026memrl},
Memp maintains evolving procedural
instructions~\citep{fang2025memp}, and ReasoningBank
extracts generalizable strategies from self-judged
successful and failed
trajectories~\citep{ouyang2025reasoningbank}.
These methods improve memory management through
task feedback, learned strategies, or reusable experience.
Our method extends guideline-based supervision
with a world model of memory dynamics that captures
how information evolves across successive updates.
We use the likelihoods of observed transitions to
adaptively weight update scores, connecting predictive
feedback with trajectory-level policy optimization.
Our memory damage check further uses dialogue evidence
to identify unsupported revisions, unjustified omissions,
and loss of essential context.

\subsection{World Models and Predictive Feedback}

World models support agent learning and decision-making
by predicting action consequences.
Dreamer 4 learns control policies within a world model
trained from offline data~\citep{hafner2025dreamer4}.
For web agents, WMA predicts salient observation
changes~\citep{chae2025wma}, while WebDreamer simulates
candidate outcomes for planning~\citep{gu2024webdreamer}.
WebEvolver combines training experience generation
with look-ahead decisions~\citep{fang2025webevolver},
and WebWorld scales learned simulation for training
and search~\citep{xiao2026webworld}.
A related line of work studies feedback for individual
decisions.
AgentPRM learns process evaluations from Monte Carlo
rollout targets~\citep{choudhury2025agentprm};
iStar derives implicit step rewards from trajectory
preferences~\citep{liu2025istar}; and SALT refines
step-level advantages through trajectory
graphs~\citep{li2026salt}.
While these world models simulate external environments
and process reward methods evaluate intermediate decisions,
our method models the evolution of the agent's own memory.
We learn from observed information-state dynamics
and use their likelihoods to modulate guideline-based
process rewards, connecting predictive modeling
with feedback for individual memory updates.
Our approach uses memory evolution as a source of
training supervision without requiring imagined
environment rollouts or return prediction.
We further combine the predictive feedback with evidence-grounded
damage assessment to distinguish predictable changes
from justified updates.

\begin{figure*}
        \centering
        \includegraphics[width=\linewidth]
        {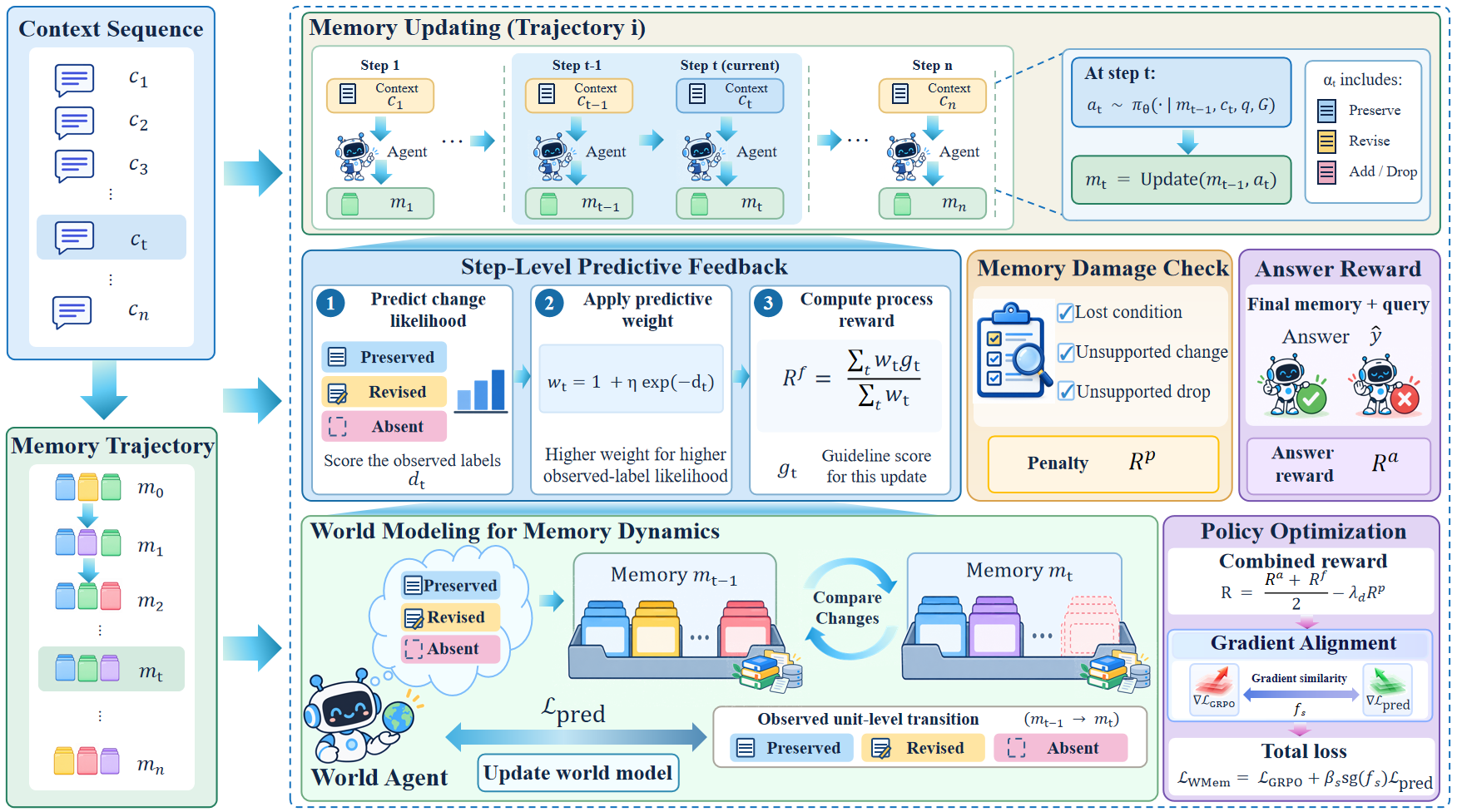}
        \caption{\textbf{Overall framework of WMem.} Given a context chunk, a query, and a fixed guideline, the policy sequentially updates memory, while the world model predicts memory-state transitions and learns from observed outcomes. Step-Level Predictive Feedback uses dynamics prediction errors to reweight guideline scores, which are combined with task rewards and evidence-based memory damage penalties to guide GRPO updates.}
        \label{framework}
\end{figure*}

\section{Method}
\label{sec:method}

We introduce WMem, a framework that learns memory-update
policies through information-state prediction and predictive
step feedback. WMem jointly optimizes a memory policy model
and a shared-parameter world model, uses observed dynamics
likelihoods to reweight memory-update scores, and incorporates
 the evidence-constrained supervision to discourage information
damage. 

\subsection{Problem Formulation}
\label{sec:problem_formulation}

\paragraph{Sequential memory updating.}
Let a training instance be
$x=(\mathcal{C},q,y^{*})$, where
$\mathcal{C}=(c_1,\ldots,c_n)$ is an interaction history
partitioned into ordered context chunks, $q$ is a downstream
query, and $y^{*}$ is its reference answer.
The agent maintains a bounded textual memory $m_t$ after
processing chunk $c_t$, starting from an empty memory
$m_0=\varnothing$.
At step $t$, policy receives update prompt with guideline $\mathcal{G}$  following MemCoE \citep{xu2026memcoe}:
\begin{equation}
    \kappa_t
    =
    \textit{Aggr}
    (\mathcal{G}, c_{t},m_{t-1})
    \label{eq:update_context}
\end{equation}
where $\textit{Aggr}(\cdot)$ means aggregation the prompt and then generates a memory-update action by the policy model $\pi_{\theta}$:
\begin{equation}
    a_t\sim\pi_\theta(\cdot\mid\kappa_t),
    \qquad
    m_t=\operatorname{Apply}(m_{t-1},a_t).
    \label{eq:memory_policy}
\end{equation}
Here, $a_t$ is a textual action that specifies a complete
rewrite of the memory, while $m_t$ denotes the resulting
memory state.
Preservation, revision, and removal are realized through
the rewritten content rather than a separate discrete
action vocabulary.
The output length is bounded by a memory-token budget.
After processing all chunks, the same language model
generates an answer:
\begin{equation}
    \hat{y}
    \sim
    \pi_\theta
    \bigl(
        \cdot\mid\operatorname{\textit{Aggr}}(q,m_n)
    \bigr).
    \label{eq:answer_policy}
\end{equation}

\paragraph{History-augmented decision process.}
We formulate the interaction as a finite-horizon,
history-augmented partially observable Markov
decision process (POMDP):
\begin{equation}
    \mathcal{M}
    =
    (\mathcal{S},\mathcal{O},\mathcal{A},P,r,\gamma),
    \label{eq:mdp}
\end{equation}
where $\mathcal{S}$ and $\mathcal{O}$ are the state
and observation spaces, $\mathcal{A}$ is the space
of generated text actions, $P$ is the transition
kernel, $r$ is the reward function and
we set $\gamma=1$ for undiscounted finite-horizon
optimization.
For a task instance $x$, the environment state is:
\begin{equation}
    s_t=(x,t,m_{0:t-1}),
    \label{eq:full_state}
\end{equation}
The policy observes the prompt $o_t=\kappa_t$
and generates an action $a_t$.
The environment applies this action to obtain $m_t$,
records the updated memory, and advances the context index:
\begin{equation}
    s_{t+1}
    =
    F(s_t,a_t)
    =
    \bigl(x,t+1,
    m_{0:t-1}\oplus m_{t}),
    \label{eq:state_transition}
\end{equation}
where $\oplus$ appends a memory state to the sequence.
The next observation $o_{t+1}$, specified by $\Omega$ from $s_{t+1}$, contains
$m_t$ and $c_{t+1}$ for another update, or $m_n$ and
the query for the final answer.
The history-augmented state satisfies the Markov property:
\begin{equation}
    P(s_{t+1}\mid s_{1:t},a_{1:t})
    =
    P(s_{t+1}\mid s_t,a_t).
    \label{eq:markov_property}
\end{equation}

An additional terminal action produces $\hat{y}$ finally.
The full state describes the environment rather than
the policy input: unread chunks and the reference
answer remain hidden.
We therefore do not assume that $m_{t-1}$ is a
sufficient statistic for the complete interaction history.
Let $\tau=(a_1,\ldots,a_n,\hat{y})$ denote a generated
action trajectory, with memory states determined
by applying its updates.
The underlying task objective is as follow:
\begin{equation}
    \max_\theta
    \mathbb{E}_{x\sim\mathcal{D}}
    \mathbb{E}_{\tau\sim\pi_\theta(\cdot\mid x)}
    \left[Y(\tau)\right],
    \label{eq:task_objective}
\end{equation}
where $Y(\tau)$
is the reference answer score in the downstream query, with $Y(\tau)\in\{0,1\}$
in our training setup.
WMem introduces additional supervision to learn
memory updates that preserve the relevant information
and incorporate new useful evidence, supporting more
accurate answers to downstream queries.



\subsection{World Modeling for Memory Dynamics}
\label{sec:memory_world_model}

\paragraph{Memory Dynamics Representation and Supervision.}
We characterize memory dynamics through the evolution
of information units across consecutive memory states.
For trajectory $\tau$, a deterministic procedure
extracts the sets of complete textual units
from the memory $m_{t}$ at the step $t$:
\begin{equation}
   \{u_{t,j}\}_{j=1}^{K_{t}}
    = \operatorname{ExtractUnits}(m_{t}),
    \label{eq:unit_selection}
\end{equation}
where $K_{t}$ is the number of extracted units.
The procedure retains intact memory lines, preserving
temporal qualifiers, contextual conditions, and negations.
Each unit may contain multiple related facts,
allowing its content to be tracked together with
the conditions under which it applies. We describe each unit's state in the subsequent
memory using three labels:
\begin{equation}
    \mathcal{Z}
    = \{\mathrm{preserved},\mathrm{revised},\mathrm{absent}\}.
    \label{eq:state_labels}
\end{equation}
A unit is \emph{preserved} if its substantive content
and qualifying conditions remain expressed, including
through paraphrase.
It is \emph{revised} if corresponding information
remains but its value, validity, temporal scope,
or contextual conditions change.
It is \emph{absent} if the subsequent memory no longer
expresses the information it conveys.
These categories capture semantic changes rather
than differences in wording alone. To obtain supervision, a frozen Qwen2.5-7B-Instruct
annotator $J_{\mathrm{state}}$ compares each unit
with the subsequent memory:
\begin{equation}
    z_{t,j}
    =
    J_{\mathrm{state}}
    (u_{t,j},m_{t}, m_{t+1}),
    \label{eq:state_annotation}
\end{equation}
where $z_{t,j}\in\mathcal{Z}$ denotes the dynamics pseudo-label of unit $j$ at step $t$.
Labels of \emph{preserved} or \emph{revised} require
supporting quotations from $m_{t+1}$.
The labels supervise learning of
memory dynamics by describing observed information
changes, without establishing whether those changes
are justified to train predictive world model.

\paragraph{Shared-Parameter World Model.}
For each annotation, we construct a prediction
prompt from prediction instruction prompt $P_{pred}$, policy prompt $\kappa_{t}$,
generated memory $m_{t}$, and extracted
unit $u_{t,j}$:
\begin{equation}
    v_{t,j}
    =
    \textit{Aggr}
    (P_{pred}, \kappa_{t}, m_{t}, u_{t,j}).
    \label{eq:prediction_input}
\end{equation}
The world model shares all language-model parameters
$\theta$ with the memory policy model.
Given $v_{t,j}$, it predicts the unit's state
after the next memory update.
The next context chunk $c_{t+1}$ and memory
$m_{t+1}$ are excluded from the prediction input,
as are annotation quotations, reference answers,
and reward values.
Thus, future information supplies the training
target but is not available to the predictor. Let $\mathcal{B}_{\mathrm{pred}}$ denote the set of
memory units examples $(v_{t,j},z_{t,j})$.
We map the three dynamics labels (preserved, revised, absent) to single tokens
using $\ell:\mathcal{Z}\rightarrow
\{\mathrm{A},\mathrm{B},\mathrm{C}\}$ and train the
model to generate label token followed by
$\mathrm{EOS}$.
Prediction objective averages negative
log-likelihood over these two target tokens:
\begin{equation}
\begin{aligned}
    \mathcal{L}_{\mathrm{pred}}
    ={}&
    -\frac{1}{2|\mathcal{B}_{\mathrm{pred}}|}
    \sum_{(v_{t,j}, z_{t,j})\in \mathcal{B}_{\mathrm{pred}}}
    \Big[
        \log \pi_\theta
        \bigl(\ell(z_{t,j})\mid v_{t,j}\bigr)
    \\
    &\qquad+
        \log \pi_\theta
        \bigl(\mathrm{EOS}\mid
        v_{t,j},\ell(z_{t,j})\bigr)
    \Big].
    \label{eq:prediction_loss}
\end{aligned}
\end{equation}
This objective couples memory-policy learning with
unit-level dynamics prediction through shared parameters,
without future-dialogue generation or
simulation of alternative update trajectories.

\subsection{Step-Level Predictive Feedback}
\label{sec:predictive_credit}

Guideline scores assess update quality, while predictive
likelihoods indicate how well the world model captures
the resulting information changes.
We combine these complementary signals through bounded
weights that adjust each update score's contribution
to the process reward, incorporating learned memory
dynamics into trajectory-level optimization of
the memory policy.


\paragraph{Predictive Likelihood.}
Before updating the policy model, we evaluate the
state labels using the previous parameter snapshot
$\theta_{\mathrm{old}}$.
We compute the mean
negative log-likelihood of their observed labels obtained by the annotaor above:
\begin{equation}
    d_{t}
    =
    -\frac{1}{|K_{t-1}|}
    \sum_{j=1}^{K_{t-1}}
    \log \pi_{\theta_{\mathrm{old}}}
    \bigl(
        \ell(z_{t-1,j})
        \mid v_{t-1,j}
    \bigr),
    \label{eq:transition_nll}
\end{equation}
where $\ell(z_{t-1,j})$ is the label token
corresponding to the annotated information state,
and $v_{t-1,j}$ is the associated prediction prompt.
Probabilities are computed over the full vocabulary
without renormalization over label tokens. We convert it into update weight:
\begin{equation}
    w_{t}=1+\eta\exp(-d_{t})
    \label{eq:predictive_weight}
\end{equation}
Here, $\exp(-d_{t})$ is the geometric mean
of the observed state-token probabilities.
The weight is attached to update $t$, which produces
the memory in which the labeled changes are observed.
The first update and updates without valid labels
receive unit weight.

\paragraph{Weighted Process Reward.}
For a trajectory with $n$ updates, each guideline
score $g_{t}$ evaluates how well the corresponding
update follows the fixed memory-management guideline,
including information retention, evidence integration,
and organizational consistency according to \citep{xu2026memcoe}.
We aggregate these scores as:
\begin{equation}
    R^{f}
    =
    \frac{
        \sum_{t=1}^{n}w_{t}g_{t}
    }{
        \sum_{t=1}^{n}w_{t}
    }.
    \label{eq:weighted_guideline}
\end{equation}
Inspired by Nemori \citep{ma2026nemori}, which
uses discrepancies between predicted content and
observed interactions to identify missing knowledge
and enrich semantic memory.
We bring this predictive perspective to policy learning:
the auxiliary objective improves the modeling of
memory dynamics, while reward weighting gives
modestly greater emphasis to update scores whose
associated changes are better captured by the model.
These mechanisms connect assessments of the
update quality with the learned dynamics of memory.


\definecolor{PersonaMemBG}{RGB}{255,229,229}
\definecolor{PrefEvalBG}{RGB}{235,235,255}
\definecolor{PersonaBenchBG}{RGB}{255,239,222}
\definecolor{OverallBG}{RGB}{229,238,247}

\definecolor{bestbg}{RGB}{255,222,222}
\definecolor{secondbg}{RGB}{222,237,252}

\providecommand{\bestscore}[1]{\cellcolor{bestbg}\textbf{#1}}
\providecommand{\secondscore}[1]{\cellcolor{secondbg}#1}

\begin{table*}[t]
\centering
\caption{\textbf{Overall comparison across eight evaluation settings.}
We report results on PersonaMem (32K/128K) (In-Domain),
PrefEval (Explicit/Implicit) (Out-of-Domain), and PersonaBench
under different noise levels (Out-of-Domain).
Higher is better. The best and second-best results in each column
are highlighted in light red and light blue, respectively.}
\label{tab:overall_comparison}

\begingroup
\setlength{\tabcolsep}{6pt}
\renewcommand{\arraystretch}{1.2}

\resizebox{\textwidth}{!}{%
\begin{tabular}{l|cc|cc|cccc|c}
\toprule

\multirow{2}{*}{\textbf{Method}}
& \multicolumn{2}{c|}{
    \cellcolor{PersonaMemBG}\textbf{PersonaMem}}
& \multicolumn{2}{c|}{
    \cellcolor{PrefEvalBG}\textbf{PrefEval}}
& \multicolumn{4}{c|}{
    \cellcolor{PersonaBenchBG}\textbf{PersonaBench (Noise Level)}}
& \cellcolor{OverallBG}
\\

& \cellcolor{PersonaMemBG}\textbf{32K}
& \cellcolor{PersonaMemBG}\textbf{128K}
& \cellcolor{PrefEvalBG}\textbf{Explicit}
& \cellcolor{PrefEvalBG}\textbf{Implicit}
& \cellcolor{PersonaBenchBG}\textbf{w/o Noise}
& \cellcolor{PersonaBenchBG}\textbf{0.3}
& \cellcolor{PersonaBenchBG}\textbf{0.5}
& \cellcolor{PersonaBenchBG}\textbf{0.7}
& \cellcolor{OverallBG}\multirow{-2}{*}{\textbf{Overall}}
\\
\midrule

Long Context
& 34.60 & 25.05
& 31.70 & 30.80
& 29.00 & 19.10 & 17.83 & 13.00
& 26.94 \\

RAG
& 47.40 & 38.90
& 47.80 & 32.40
& 29.09 & 28.16 & 24.31 & 23.00
& 36.46 \\

\midrule

Mem0
& 46.02 & 39.67
& 57.60 & 46.40
& 17.60 & 19.75 & 19.22 & 17.80
& 37.81 \\

A-Mem
& 47.06 & 38.22
& 62.30 & 52.80
& 30.32 & 28.56 & 25.19 & 24.45
& 42.44 \\

LightMem
& 48.79 & 39.93
& 64.20 & 54.80
& 19.08 & 18.74 & 19.65 & 17.80
& 40.89 \\

\midrule

MemAgent
& 56.40 & 48.40
& 75.30 & 65.50
& 20.10 & 19.29 & 16.57 & 16.20
& 46.95 \\

Mem-$\alpha$
& 55.36 & 47.45
& 73.30 & 63.40
& 19.37 & 17.68 & 16.4 & 12.33
& 45.40 \\

MemSkill
& 54.33
& 49.25
& 80.20
& 68.20
& 32.67
& 29.53
& 26.31
& \bestscore{25.87}
& 51.53 \\

MemCoE
& \secondscore{62.88}
& \secondscore{52.69}
& \secondscore{81.60}
& \secondscore{76.50}
& \secondscore{32.69}
& \secondscore{29.96}
& \secondscore{26.83}
& 25.56
& \secondscore{55.20} \\

\textbf{WMem (Ours)}
& \bestscore{71.11}
& \bestscore{58.64}
& \bestscore{85.30}
& \bestscore{81.30}
& \bestscore{33.87}
& \bestscore{30.27}
& \bestscore{27.70}
& \secondscore{25.70}
& \bestscore{59.19} \\

\bottomrule
\end{tabular}%
}

\endgroup
\end{table*}

\subsection{Memory Damage Check}
\label{sec:evidence_damage}

Predictive likelihood does not establish whether
a memory change is justified, while overall
guideline scores may overlook localized information
damage.
We therefore introduce an evidence-grounded penalty
that explicitly accounts for unsupported changes,
unjustified omissions, and loss of essential conditions.

For update $t$, we obtain
$K_{t-1}$ memory units from the previous memory
$m_{t-1}$, independently of the generated update.
A frozen evaluator examines these units using
current chunk, retrieved excerpts from already-consumed
dialogue, and the updated memory.

We consider three damage categories:
\emph{unsupported change},
\emph{unsupported drop}, and
\emph{lost condition}.
The last category captures removal of a material
temporal, contextual, or negation constraint.
Valid revisions, reasonable compression, and justified
removal receive no penalty. For a selected unit $j$, We define:
\begin{equation}
    D_{t,j}
    =
    \mathbb{I}
    \left[
        \text{a damage annotation passes evidence checks}
    \right].
    \label{eq:unit_damage}
\end{equation}
An annotation must cite an exact quotation
from a supplied raw-dialogue source.
Claims of changed content additionally require a
quotation from the updated memory.
Ambiguous, malformed, and unsupported annotations
receive $D_{t,j}=0$. The trajectory-level damage score is computed as follow:
\begin{equation}
    R^{p}
    =
    \frac{
        \sum_{t=1}^{n}
        \sum_{j=1}^{K_{t-1}}
        D_{t,j}
    }{
          \sum^{n}_{t=1}K_{t-1}
    }.
    \label{eq:trajectory_damage}
\end{equation}
All selected units remain in the denominator,
including those with uncertain or rejected annotations.
Thus, abstention does not increase the contribution
of other penalties. Evidence checks use only already-consumed dialogue
and do not access unread chunks or reference answers.
They provide conservative process supervision,
while remaining subject to frozen evaluator's
semantic judgment.

\subsection{Trajectory Reward and Policy Optimization}
\label{sec:policy_optimization}

\paragraph{Combined trajectory reward.}

For each trajectory $\tau$, we combine the answer
reward $R^{a}$, predictively weighted guideline
reward $R^{f}$, and memory damage penalty $R^{p}$
to obtain the trajectory reward:
\begin{equation}
    R
    =
    \frac{1}{2}
    \left(R^{a}+R^{f}\right)
    -
    \lambda_{d} R^{p},
    \label{eq:combined_reward}
\end{equation}
where $R^{a}
\in\{0,1\}$ indicates whether the final answer $y^{*}$ is
equal to $\hat{y}$, providing task-level feedback on the
usefulness of the maintained memory.
The nonnegative penalty $R^{p}$ measures
evidence-supported memory damage, and $\lambda_d$
controls its contribution.
The resulting rewards
$\mathbf{R}=(R_1,\ldots, R_{i},\ldots, R_G)$ are used for
group-relative advantage estimation.

 
\paragraph{Group-relative policy optimization.}
For each training instance, we sample $G$ trajectories
using the policy $\pi_{\theta_{\mathrm{old}}}$.
We compute each trajectory's advantage by centering
its reward within the group:
\begin{equation}
    \bar{R}
    =
    \frac{1}{G}\sum_{i=1}^{
    G
    }R_i,
    \qquad
    A_i=R_i-\bar{R}.
    \label{eq:group_advantage}
\end{equation}
We use $G=2$ and omit normalization by the
within-group standard deviation.
The resulting group-relative advantage $A_i$ is
assigned to all generated memory and answer tokens
in trajectory $\tau_i$ and used in the clipped
GRPO objective.
Therefore, the predictive feedback and the memory damage
assessment guide policy learning through the
all combined trajectory reward.


\paragraph{GRPO-based policy optimization.}
We optimize the policy using group-relative advantages
$A_i$ computed from the combined trajectory rewards.
For each generated token $a_{i,k}$ with context
$h_{i,k}$, let
$\rho_{i,k}(\theta)=
\pi_\theta(a_{i,k}\mid h_{i,k})/
\pi_{\theta_{\mathrm{old}}}(a_{i,k}\mid h_{i,k})$
denote the policy probability ratio.
The policy objective is:
\begin{equation}
\begin{aligned}
\mathcal{L}_{\mathrm{GRPO}}
= {}&
-\frac{1}{Z}\sum_{i,k}M_{i,k}
\min\Bigl(
\rho_{i,k}(\theta)A_i,\\
&\qquad
\operatorname{clip}\bigl(
\rho_{i,k}(\theta),1-\epsilon,1+\epsilon
\bigr)A_i
\Bigr)
+\lambda_{\mathrm{KL}}\mathcal{L}_{\mathrm{KL}}
\end{aligned}
\label{eq:main_loss}
\end{equation}
where $M_{i,k}$ masks valid generated tokens,
$Z=\sum_{i,k}M_{i,k}$, and
$\mathcal{L}_{\mathrm{KL}}$ regularizes the policy
against a frozen reference model.
We set $\epsilon=0.2$ and
$\lambda_{\mathrm{KL}}=10^{-3}$.
Prompt and padding tokens are excluded, and
advantages are treated as constants.
Our implementation additionally uses dual clipping,
as detailed in the appendix.

\subsection{Gradient Predictive Learning Coefficient}
\label{sec:gradient_alignment}

The auxiliary predictor and the memory policy share
parameters, so their gradients may favor different
updates. We regulate the strength of prediction
learning using their gradient alignment:
\begin{equation}
    \mathbf{g}_{\mathrm{GRPO}}
    =
    \nabla_\theta\mathcal{L}_{\mathrm{GRPO}},
    \qquad
    \mathbf{g}_{\mathrm{pred}}
    =
    \nabla_\theta\mathcal{L}_{\mathrm{pred}}
    \label{eq:task_gradients}.
\end{equation}
Their cosine similarity is computed as:
\begin{equation}
    c_s
    =
    \frac{
        \langle
            \mathbf{g}_{\mathrm{GRPO}},
            \mathbf{g}_{\mathrm{pred}}
        \rangle
    }{
        \|\mathbf{g}_{\mathrm{GRPO}}\|_2
        \|\mathbf{g}_{\mathrm{pred}}\|_2
    }.
    \label{eq:gradient_cosine}
\end{equation}
We set $c_s=0$ when either norm is zero or the
similarity cannot be computed reliably.
The alignment factor is:
\begin{equation}
    f_s
    =
    1+\xi\operatorname{clip}(c_s,-1,1),
    \label{eq:alignment_factor}
\end{equation}
where $\xi=0.1$.
This bounds the modulation to $f_s\in[0.9,1.1]$.

\begin{figure*}[t]
    \centering
    \begin{subfigure}[t]{0.49\textwidth}
        \centering
        \includegraphics[width=\linewidth]
        {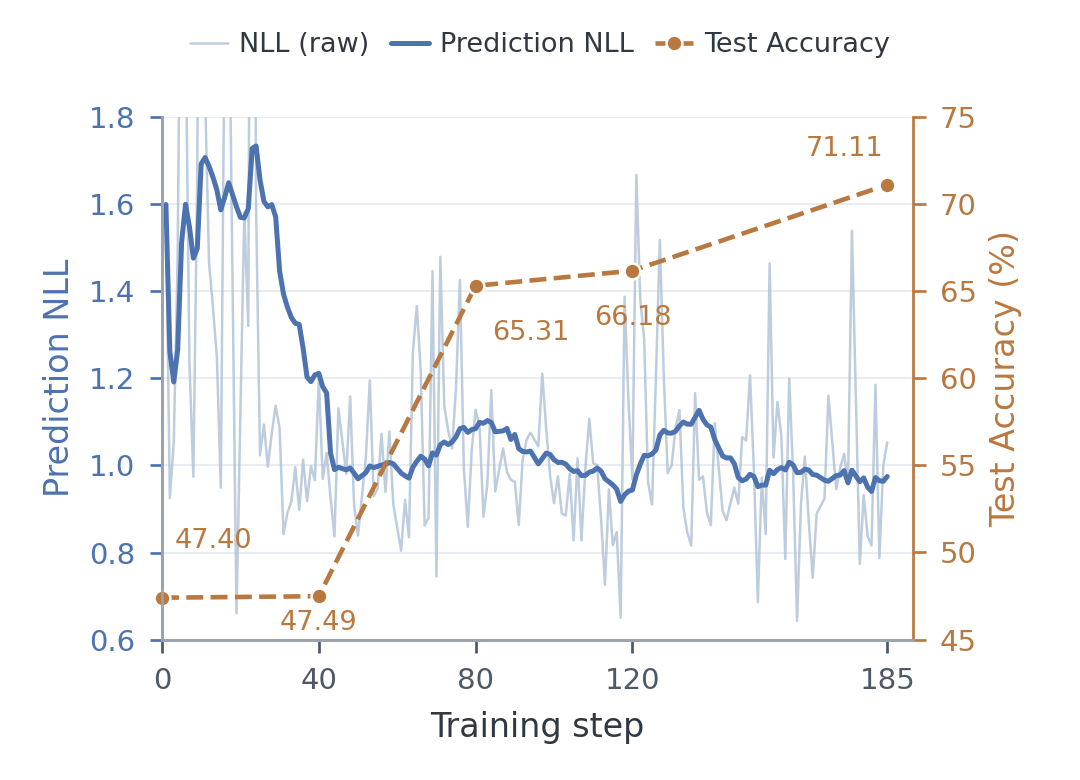}
        \caption{Prediction error and test accuracy.}
        \label{fig:prediction-test}
    \end{subfigure}
    \hfill
    \begin{subfigure}[t]{0.49\textwidth}
        \centering
        \includegraphics[width=\linewidth]
        {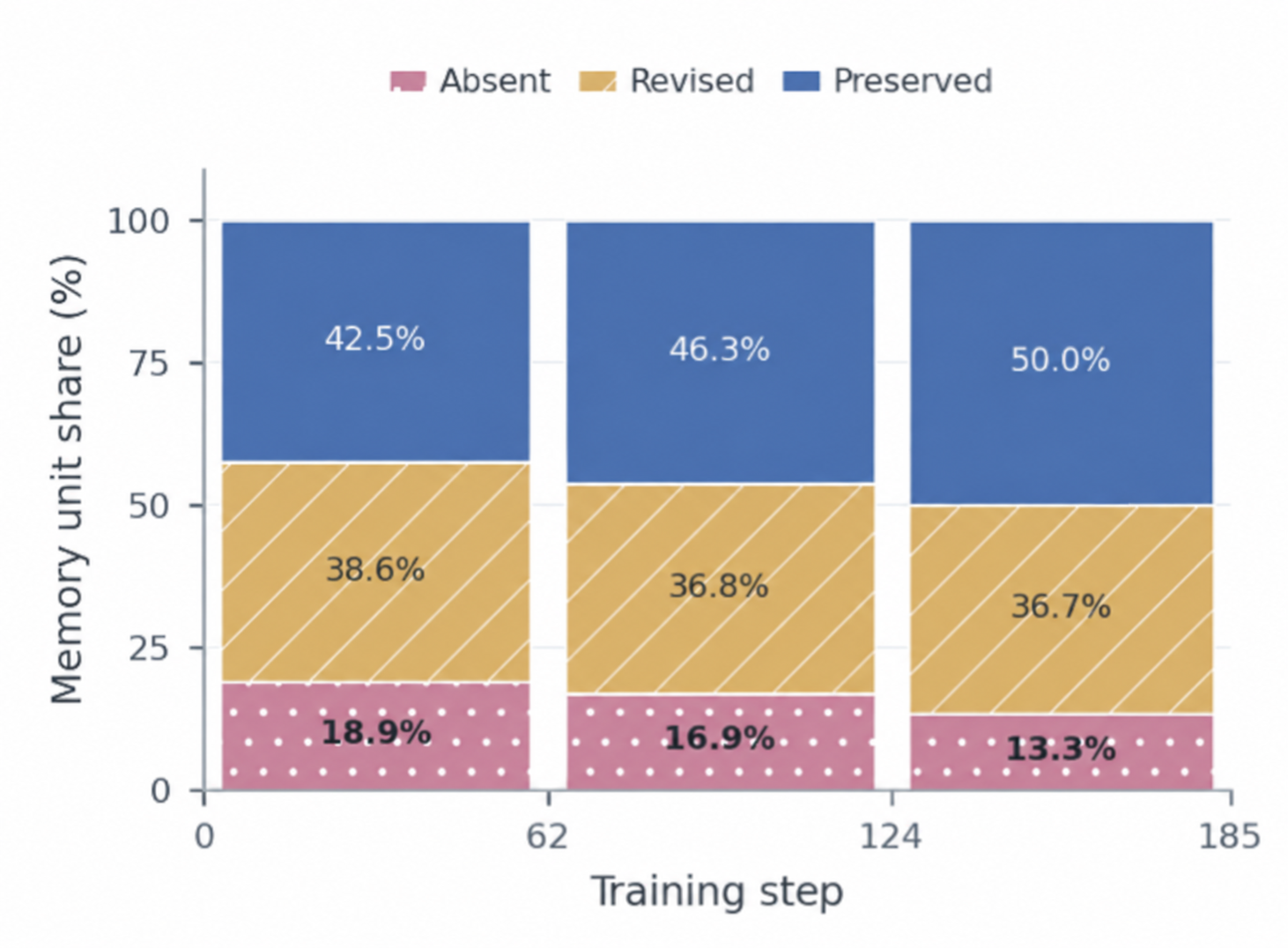}
        \caption{Memory state composition.}
        \label{fig:memory-composition}
    \end{subfigure}
    \caption{Training dynamics of WMem methods on PersonMem 32K.
    Left: prediction NLL and test accuracy.
    Right: proportions of preserved, revised, and absent
    memory units, aggregated within each training interval.}
    \label{fig:method13-training}
\end{figure*}


After obtaining the total reward including task reward, stepped guideline reward and the penalty reward, we can get the per-step objective with the world agent predictive loss:
\begin{equation}
    \mathcal{L}_{\mathrm{WMem}}
    =
    \mathcal{L}_{\mathrm{GRPO}}
    +
    \beta_s\operatorname{sg}(f_s)
    \mathcal{L}_{\mathrm{pred}}.
    \label{eq:total_loss}
\end{equation}
where the $\beta(s)$ we set here is 0.002. The alignment factor is computed from the current
gradients and held fixed during the update, without
second-order differentiation.
We then apply the existing gradient clipping and
AdamW optimizer.




\section{Experiments}

\subsection{Experiment Setup}
\label{sec:experimental_settings}

\subsubsection{Benchmarks}
Following the general experimental setup of
MemCoE~\citep{xu2026memcoe}, we evaluate WMem on three
benchmarks covering different context lengths, query
types, and noise levels.
\textbf{PersonaMem}~\citep{jiang2025know}
serves as our in-domain benchmark,
with 32K and 128K settings that assess memory-based
question answering over extended dialogue histories.
We train on 300 questions from PersonaMem-32K and
evaluate on held-out questions.
\textbf{PrefEval}~\citep{zhao2025preferences}
serves as an out-of-domain benchmark
with explicit and implicit preference settings,
testing whether the maintained memory captures
information that is directly stated or must be
inferred from interactions.
\textbf{PersonaBench}~\citep{tan2025personabench}
further evaluates out-of-domain performance
under noisy evidence.
We consider a noise-free setting and noise levels
of 0.3, 0.5, and 0.7 to examine robustness
as noise increases.
Together, these benchmarks provide eight evaluation
settings for assessing the effectiveness,
generalization, and noise robustness of memory updates.

\subsubsection{Baselines}
We consider three categories of baselines to evaluate our
framework against different types of memory management paradigms.
\textbf{Context-based methods} include
\textit{Long Context}~\cite{liu2024lost}, which directly answers
queries using the dialogue history within the available context
window, relying on the model's contextual understanding, and
\textit{RAG}, which retrieves relevant dialogue segments based
on the current query to support answer generation.
\textbf{Memory-system baselines} explicitly maintain external
memories for long-term information retention, including
\textit{Mem0}~\citep{chhikara2025mem0}, which extracts,
consolidates, and retrieves salient conversational
information;
\textit{A-MEM}~\citep{xu2025amem}, which organizes
memories into interconnected notes with dynamic linking
and revision as new information becomes available; and
\textit{LightMem}~\citep{fang2025lightmem}, which combines
sensory filtering, topic-aware consolidation, and
offline long-term memory updates for efficient memory management.
\textbf{Reinforcement Learning-based memory methods} optimize memory
management policies through learning-based objectives, including
\textit{MemAgent}~\citep{yu2025memagent}, which learns
sequential memory compression and updating through
reinforcement learning;
\textit{Mem-$\alpha$}~\citep{wang2025memalpha}, which
learns adaptive memory construction and tool-based operations
using downstream question-answering feedback;
\textit{MemSkill}~\citep{zhang2026memskill}, which
learns to select reusable memory skills and progressively refines
the skill set by reviewing difficult cases; and
\textit{MemCoE}~\citep{xu2026memcoe}, which combines the
memory guideline induction with guideline-aligned
reinforcement learning to guide the memory update decisions.


\subsubsection{Implementation Details}
\label{sec:implementation}

Following settings of MemCoE, we initialize WMem with Qwen2.5-7B-Instruct \cite{yang2024qwen25}, sharing parameters between the memory update policy and the world model. Additionally, we employ a separate, frozen Qwen2.5-7B-Instruct model as the memory dynamics annotator to identify memory state transitions, whose parameters remain fixed throughout training. All the Experiments were conducted on 8 NVIDIA H200 GPUs.  For retrieval, we adopt all-MiniLM-L6-v2 \citep{wang2020minilm} and retrieve the top-10 candidate. We use AdamW with a learning rate of $10^{-6}$ and 20 warmup steps. The predictive loss coefficient is set as $2 \times 10^{-3}$. Predictive reward weighting and gradient-alignment modulation both use a scale of 0.1. The final reward equally weights task success and the prediction-weighted guideline score, with a damage penalty coefficient of 0.01. More detailed information about the benchmarks and training parameters are provided in supplemental materials.

\subsection{Main Results}

\begin{table}[t]
    \centering
    \caption{Ablation study of each components of WMem on PersonaMem and PrefEval benchmarks.}
    \label{tab:ablation}
    \renewcommand{\arraystretch}{1.2}
    \setlength{\tabcolsep}{4pt}

    \begin{tabularx}{\columnwidth}{
        l|
        *{2}{>{\centering\arraybackslash}X}|
        *{2}{>{\centering\arraybackslash}X}
    }
        \toprule
        \multirow{2}{*}{\textbf{Setting}}
        & \multicolumn{2}{c|}{\textbf{PersonaMem}}
        & \multicolumn{2}{c}{\textbf{PrefEval}} \\
        & \textbf{32K} & \textbf{128K}
        & \textbf{Explicit} & \textbf{Implicit} \\
        \midrule

        \textbf{WMem}
        & \textbf{71.11} & \textbf{58.64}
        & \textbf{85.30} & \textbf{81.30} \\

        w/o Penalty
        & 69.46 & 57.06 & 80.40 & 77.60  \\

        w/o EOS
        & 68.51 & 56.44  & 79.50 & 74.20 \\

        w/o SF (Step Feedback) 
        & 68.25 & 56.11 & 77.30 & 73.70 \\

        w/o $L_{pred}$
        & 68.17 & 54.64 & 76.10 & 73.50 \\

        w/o Penalty \& SF
        & 68.08 & 55.99 & 75.80 & 71.40 \\

        w/o Coefficient
        & 66.18 & 55.01 & 74.60 & 68.80 \\

        \bottomrule
    \end{tabularx}
\end{table}

\begin{table}[t]
    \centering
    \caption{Effectiveness of the world agent's prediction horizon.}
    \label{tab:prediction_steps}
    \renewcommand{\arraystretch}{1.2}
    \setlength{\tabcolsep}{0pt}
    \begin{tabular}{
        >{\raggedright\arraybackslash}
        p{\dimexpr0.32\columnwidth-\arrayrulewidth\relax}|
        >{\centering\arraybackslash}p{0.17\columnwidth}
        >{\centering\arraybackslash}
        p{\dimexpr0.17\columnwidth-\arrayrulewidth\relax}|
        *{2}{>{\centering\arraybackslash}p{0.17\columnwidth}}
    }
        \toprule
        & \multicolumn{2}{c|}{\textbf{PersonaMem}}
        & \multicolumn{2}{c}{\textbf{PrefEval}} \\
        \textbf{Setting}
        & \textbf{32K} & \textbf{128K}
        & \textbf{Explicit} & \textbf{Implicit} \\
        \midrule
        \textbf{1-Step Horizon}
        & \textbf{71.11} & \textbf{58.64}
        & \textbf{85.30} & \textbf{81.30} \\
        2-Step Horizon
        & 66.18 & 56.44 & 78.50 & 77.70 \\
        3-Step Horizon
        & 65.31 & 49.21 & 78.00 & 77.50 \\
        \bottomrule
    \end{tabular}
\end{table}

\begin{figure*}
            \centering
        \includegraphics[width=\linewidth]
        {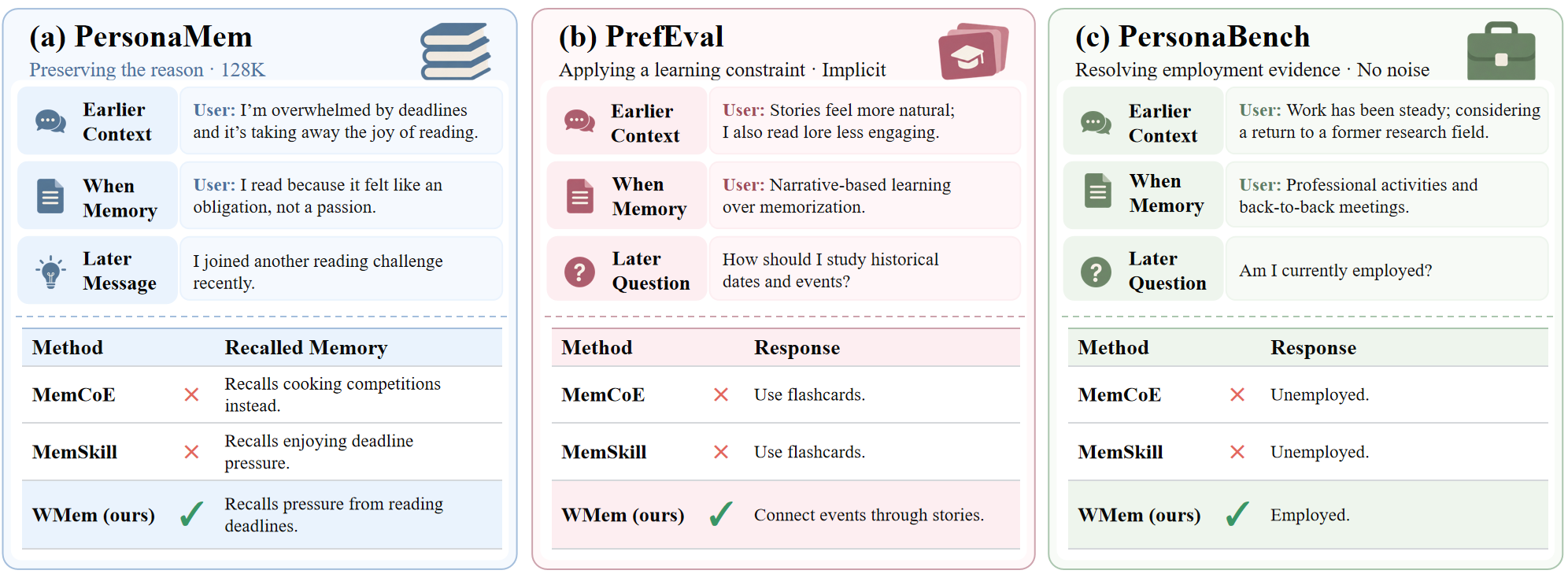}
       \caption{\textbf{Case studies across three benchmarks.}
WMem recalls reading-related pressure (left), applies a narrative-learning preference (middle), and correctly identifies employment status (right).
Excerpts are condensed for readability.}
        \label{casestudy}
\end{figure*}

\begin{figure}[t]
    \centering
    \includegraphics[width=\linewidth]{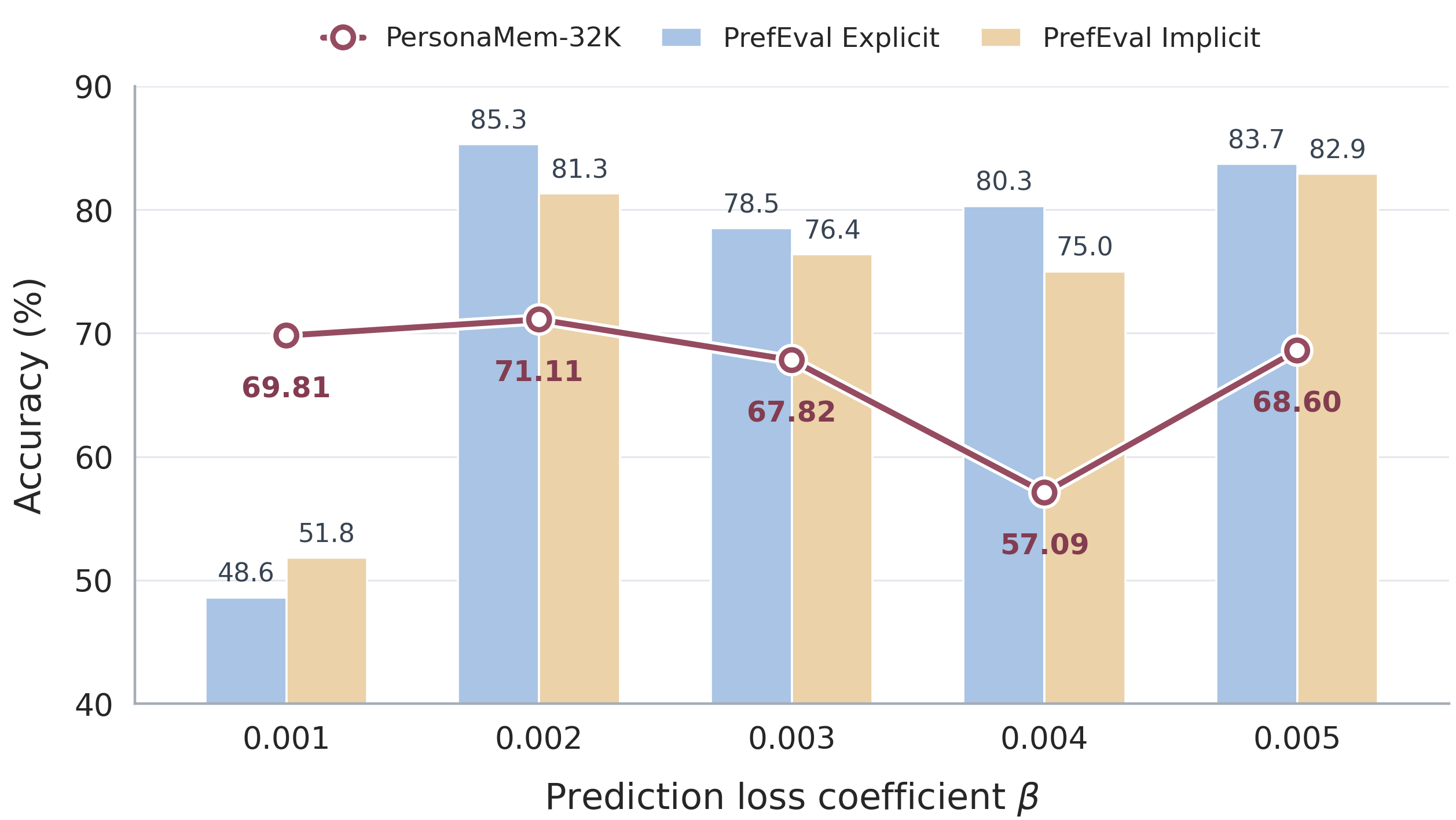}
    \caption{Effect of the prediction loss coefficient $\beta$ on PersonaMem-32K (line) and PrefEval Explicit/Implicit (bars).}
    \label{compare}
\end{figure}

\paragraph{\textbf{Overall performance.}}
Table~\ref{tab:overall_comparison} shows that WMem
achieves the best results in seven of the eight
evaluation settings, with an overall score of
59.19 compared with 55.20 for the strongest
baseline, MemCoE.
On PersonaMem, WMem achieves 71.11 and 58.64
in the 32K and 128K settings, respectively,
exceeding MemCoE by 8.23 and 5.95 percentage points.
These improvements support the effectiveness of
the learned memory-update policy across both
context lengths.
Nevertheless, performance decreases in the 128K
setting for all methods, highlighting the difficulty
of maintaining useful information over longer
interaction histories.

\paragraph{\textbf{Out-of-domain generalization.}}
WMem also improves performance on PrefEval and
PersonaBench without benchmark-specific policy training.
On PrefEval, it achieves 85.30 and 81.30 in the
explicit and implicit settings, surpassing MemCoE
by 3.70 and 4.80 percentage points, respectively.
The improvement in the implicit setting suggests
that the learned policy can retain evidence useful
for responding to preferences that are not
directly stated.
On PersonaBench, WMem achieves the highest scores
without noise and at noise levels of 0.3 and 0.5,
outperforming MemCoE by 1.18, 0.31, and 0.87
percentage points, respectively.
At a noise level of 0.7, WMem scores 25.70,
slightly below MemSkill's 25.87 but above
MemCoE's 25.56.
These results indicate generalization across
benchmarks, although the smaller gains on
PersonaBench suggest that noisy evidence remains
a substantial challenge.

\paragraph{\textbf{Training dynamics and memory quality.}}
Figure~\ref{fig:method13-training} provides insight
into the learning process.
Prediction NLL decreases from its initially high
level and subsequently fluctuates around 1.0,
while test accuracy increases from 47.40\% to 71.11\%.
Meanwhile, the proportion of preserved memory units
rises from 42.5\% to 50.0\%, and the absent proportion
decreases from 18.9\% to 13.3\%.
The revised proportion changes from 38.6\% to 36.7\%,
indicating continued revision alongside increased
retention.
These trends are consistent with improved memory
maintenance, although the state proportions alone
do not establish the correctness of individual updates.
The cases in Figure~\ref{casestudy} illustrate
how retained information supports subsequent answers.
On PersonaMem, WMem preserves the pressure associated
with reading deadlines and recalls it when reading
challenges are mentioned again.
MemCoE instead refers to cooking competitions,
while MemSkill incorrectly recalls enjoyment of
deadline pressure.
On PrefEval, WMem retains the preference for
narrative-based learning and recommends connecting
historical events through stories; both baselines
recommend less-preferred flashcards.
On PersonaBench, WMem retains evidence of ongoing
professional activity and correctly identifies
the user as employed, whereas MemCoE answers
``unemployed.''

\subsection{Ablation Study}

\paragraph{\textbf{Effect of individual components.}}
Table~\ref{tab:ablation} shows that all ablated variants
underperform full WMem across all four settings.
Removing predictive step feedback reduces accuracy
by 2.86 and 2.53 percentage points on PersonaMem-32K
and 128K, respectively, and by 8.00 and 7.60 points
on PrefEval Explicit and Implicit.
Removing $\mathcal{L}_{\mathrm{pred}}$ lowers PrefEval
Explicit and Implicit accuracy by 9.20 and 7.80 points,
respectively, supporting the value of predictive supervision
for out-of-domain performance.
Removing the penalty or EOS also consistently reduces
accuracy across benchmarks.
Jointly removing the penalty and step feedback performs
worse than removing either component alone across all
four settings and removing the coefficient yields the lowest
accuracy.

\paragraph{\textbf{Prediction horizon and prediction loss coefficient.}}
As shown in Table~\ref{tab:prediction_steps},
one-step prediction performs best across all settings.
Extending the horizon to three steps reduces
PersonaMem-128K accuracy from 58.64\% to 49.21\%,
suggesting greater uncertainty over longer horizons.
Intervening updates may make longer-range targets
less informative for the current update, favoring
supervision of immediate memory changes.
Figure~\ref{compare} shows that $\beta=0.002$
performs best on PersonaMem-32K and PrefEval Explicit,
while $\beta=0.001$ substantially degrades
PrefEval performance.
Increasing $\beta$ does not consistently improve accuracy:
although $\beta=0.005$ performs better on PrefEval
Implicit, it yields lower accuracy on other settings.
Results support $\beta=0.002$ as a balanced choice.

\section{Conclusion}
We presented WMem, a framework for learning memory
updates through world models and step-level predictive
feedback.
WMem models information dynamics and uses predictive
likelihoods to weight update-level guideline scores,
complementing task rewards and an evidence-grounded
memory damage penalty while keeping the guideline fixed.
Experiments on PersonaMem, PrefEval, and PersonaBench
demonstrate improved overall performance across
in-domain and out-of-domain settings.
Ablation studies support the contributions of
predictive supervision and reward weighting.
These findings suggest that modeling memory dynamics
can guide policies to preserve relevant information
and incorporate new evidence for reasoning over
extended interactions.



\bibliographystyle{ACM-Reference-Format}
\bibliography{main}


%
\clearpage

\appendix


\definecolor{blockblue}{RGB}{229,238,247}
\definecolor{blockpurple}{RGB}{235,235,255}
\definecolor{bestbg}{RGB}{255,222,222}
\definecolor{secondbg}{RGB}{222,237,252}

\providecommand{\bestscore}[1]{\cellcolor{bestbg}\textbf{#1}}
\providecommand{\secondscore}[1]{\cellcolor{secondbg}#1}

\definecolor{explicitblue}{RGB}{229,238,247}
\definecolor{implicitpurple}{RGB}{235,235,255}
\definecolor{leftblock}{RGB}{229,238,247}
\definecolor{rightblock}{RGB}{235,235,255}
\definecolor{promptgreen}{RGB}{0,128,0}
\definecolor{promptbackground}{RGB}{245,251,245}

\twocolumn[{
\begin{minipage}{\textwidth}
\centering
\captionof{table}{Category-wise accuracy (\%) on PersonaMem under
32K and 128K interaction histories.
``--'' indicates the category or result is unavailable.
Within each setting, the best and second-best results are
highlighted in light red and light blue, respectively.}
\label{tab:personamem_category}

\renewcommand{\arraystretch}{1}
\setlength{\tabcolsep}{6pt}

\resizebox{\textwidth}{!}{%
\begin{tabular}{l|ccccccc|c}
\toprule
\textbf{Method}
& \shortstack{\textbf{Recall}\\\textbf{facts}}
& \shortstack{\textbf{Suggest}\\\textbf{ideas}}
& \shortstack{\textbf{Latest}\\\textbf{prefs}}
& \shortstack{\textbf{Prefs}\\\textbf{evolve}}
& \shortstack{\textbf{Update}\\\textbf{reasons}}
& \shortstack{\textbf{Aligned}\\\textbf{recs}}
& \shortstack{\textbf{New}\\\textbf{Scenarios}}
& \textbf{Overall}
\\
\midrule

\multicolumn{9}{c}{\cellcolor{blockblue}
\textbf{\textit{$\blacktriangledown$ 32K memory corpus data}}}
\\
\midrule

Long Context
& 37.04 & 20.75 & -- & 37.68
& 56.00 & 37.04 & 9.70 & 34.60 \\

RAG
& 46.30 & 18.87 & -- & 59.42
& 80.00 & 37.04 & 29.03 & 47.40 \\

Mem0
& 37.29 & 16.98 & -- & 52.17
& 70.00 & 62.96 & 45.16 & 46.02 \\

A-Mem
& 45.76 & 22.64 & -- & 57.97
& 78.00 & 44.44 & 19.35 & 47.06 \\

LightMem
& 55.93 & 20.75 & -- & 57.97
& 70.00 & 48.15 & 29.03 & 48.79 \\

MemAgent
& 55.93 & 16.98 & -- & \secondscore{69.57}
& 76.00 & 55.56 & 54.84 & 56.40 \\

Mem-$\alpha$
& 57.63 & 15.09 & -- & 66.67
& 76.00 & 66.67 & 54.84 & 55.36 \\

MemSkill
& 55.93
& 22.64
& --
& 55.07
& 76.00
& 62.96
& \secondscore{61.29}
& 54.33 \\

MemCoE
& \secondscore{68.64}
& \secondscore{33.02}
& --
& 63.41
& \secondscore{82.50}
& \bestscore{75.00}
& 59.68
& \secondscore{62.88} \\

\textbf{WMem (Ours)}
& \bestscore{81.36}
& \bestscore{44.81}
& --
& \bestscore{71.38}
& \bestscore{88.00}
& \secondscore{67.59}
& \bestscore{71.77}
& \bestscore{71.11} \\

\midrule

\multicolumn{9}{c}{\cellcolor{blockpurple}
\textbf{\textit{$\blacktriangledown$ 128K memory corpus data}}}
\\
\midrule

Long Context
& 18.12
& 23.36
& 20.62
& 38.62
& 39.77
& 21.70
& 16.99
& 25.05 \\

RAG
& 54.37
& 19.67
& 36.45
& 52.69
& 58.71
& 41.94
& 29.61
& 38.90 \\

Mem0
& \secondscore{56.25}
& 20.49
& 37.77
& 55.09
& 57.20
& 41.64
& 29.13
& 39.67 \\

A-Mem
& 36.88
& 16.19
& 38.73
& 59.88
& 62.50
& 35.19
& 28.16
& 38.22 \\

LightMem
& 40.00
& 17.42
& 42.33
& 61.08
& 64.02
& 35.48
& 25.73
& 39.93 \\

MemAgent
& 50.88
& 20.08
& 49.08
& 64.34
& 63.57
& 53.87
& \secondscore{55.87}
& 48.40 \\

Mem-$\alpha$
& 48.54
& 19.69
& 50.81
& 65.40
& 59.85
& \bestscore{61.32}
& 33.33
& 47.45 \\

MemSkill
& \bestscore{61.99}
& 19.88
& 54.04
& 54.55
& \secondscore{75.84}
& \secondscore{54.44}
& 40.38
& 49.25 \\

MemCoE
& 42.11
& 18.53
& \secondscore{65.59}
& \secondscore{67.74}
& 67.66
& 49.57
& 53.99
& \secondscore{52.69} \\

\textbf{WMem (Ours)}
& 54.97
& \bestscore{33.01}
& \bestscore{68.13}
& \bestscore{69.50}
& \bestscore{79.55}
& 48.42
& \bestscore{58.22}
& \bestscore{58.64} \\

\bottomrule
\end{tabular}%
}
\end{minipage}
\vspace{1em}
}]

\section{Detailed Benchmark Results}
\label{sec:detailed_results}

\paragraph{PersonaMem.}
Table~\ref{tab:personamem_category} shows that WMem
leads in five of six available categories under
32K histories and five of seven under 128K histories.
In the 32K setting, it achieves 81.36 on
\textit{Recall Facts} and 88.00 on
\textit{Update Reasons}, exceeding MemCoE by
12.72 and 5.50 percentage points, respectively.
Under 128K histories, WMem maintains advantages
in tracking preference changes and preserving
update reasons, although it trails the strongest
baselines in factual recall and aligned recommendations.

\paragraph{PrefEval.}
Table~\ref{tab:prefeval_detailed} shows that WMem
achieves the best or tied-best results across all
seven explicit-preference domains and leads in
six of seven implicit-preference domains.
Notable gains over MemCoE include 7.90 percentage
points in explicit \textit{Shopping} and 9.81 points
in implicit \textit{Travel}.
These improvements suggest that generalization
extends across diverse topics, although WMem
remains below MemCoE in the implicit \textit{Pet}
domain.

\paragraph{PersonaBench.}
Table~\ref{tab:personabench_detailed} shows that WMem
achieves the highest overall F1 without noise
and at noise ratios of 0.3 and 0.5.
At a noise ratio of 0.5, it leads in
\textit{Basic Info}, \textit{Pref.\ (Easy)}, and
\textit{Social}, but remains weaker on
\textit{Pref.\ (Hard)}.
At a noise ratio of 0.7, WMem achieves an overall
score of 25.70, slightly below MemSkill's 25.87.
These results indicate competitive performance
under noisy conditions, while highlighting
variation in robustness across information categories.

\section{Supplementary Prompt Templates}
\label{app:prompt_templates}

\begin{tcolorbox}[
    enhanced,
    breakable,
    title={WMem --- Memory-Update Policy Prompt},
    colback=promptbackground,
    colframe=promptgreen,
    colbacktitle=promptgreen,
    coltitle=white,
    fonttitle=\bfseries,
    fontupper=\small,
    boxrule=0.7pt,
    arc=2.5mm,
    left=2.5mm,
    right=2.5mm,
    top=2mm,
    bottom=2mm
]
You are given a question with options, some new memory,
and previous user memory. Read the new section and update
the user memory by prioritizing recent and relevant
information that aligns with the user's preferences
and experiences.

\medskip
\texttt{<memory>}  
\textit{\{Previous memory $m_{t-1}$\}}
\texttt{</memory>}

\medskip
\texttt{<section>} \textit{\{Current context chunk $c_t$\}}
\texttt{</section>}

\medskip
\textbf{Update rules:} 

\textit{\{Fixed memory-management guideline $\mathcal{G}$\}}
\end{tcolorbox}


\twocolumn[{
\begin{minipage}{\textwidth}
\centering
\captionof{table}{Domain-wise accuracy (\%) on PrefEval multiple-choice
classification under explicit and implicit preferences.
Within each setting, the best and second-best results are
highlighted in light red and light blue, respectively.}
\label{tab:prefeval_domain}
\label{tab:prefeval_detailed}

\renewcommand{\arraystretch}{1.0}
\setlength{\tabcolsep}{6pt}

\resizebox{\textwidth}{!}{%
\begin{tabular}{l|ccccccc|c}
\toprule
\textbf{Method}
& \textbf{Travel}
& \textbf{Entertain}
& \textbf{Lifestyle}
& \textbf{Shop}
& \textbf{Education}
& \textbf{Professional}
& \textbf{Pet}
& \textbf{Overall}
\\
\midrule

\multicolumn{9}{c}{\cellcolor{explicitblue}
\textbf{\textit{$\blacktriangledown$ Explicit Memory}}}
\\
\midrule

Long Context
& 31.78 & 30.77 & 32.70 & 30.00
& 32.94 & 27.78 & 39.53 & 31.70 \\

RAG
& 45.33 & 52.04 & 47.87 & 45.79
& 42.35 & 58.33 & 48.84 & 47.80 \\

Mem0
& 58.41 & 61.99 & 57.35 & 53.68
& 54.12 & 69.44 & 46.51 & 57.60 \\

A-Mem
& 62.15 & 69.23 & 60.19 & 61.05
& 54.12 & 66.67 & 55.81 & 62.30 \\

LightMem
& 65.42 & 68.33 & 65.40 & 62.11
& 56.47 & 66.67 & 53.49 & 64.20 \\

MemAgent
& 77.10
& \bestscore{81.00}
& 73.46
& 74.21
& 69.41
& 77.78
& 60.47
& 75.30 \\

Mem-$\alpha$
& 78.97
& 72.40
& 72.03
& 75.79
& 62.35
& 77.78
& 62.79
& 73.30 \\

MemSkill
& 75.70
& \bestscore{81.00}
& 74.41
& \secondscore{86.84}
& \secondscore{85.88}
& 88.89
& 79.07
& 80.20 \\

MemCoE
& \secondscore{81.78}
& 76.47
& \secondscore{82.46}
& 82.63
& \secondscore{85.88}
& \secondscore{91.67}
& \bestscore{81.40}
& \secondscore{81.60} \\

\textbf{WMem (Ours)}
& \bestscore{82.71}
& \bestscore{81.00}
& \bestscore{83.89}
& \bestscore{90.53}
& \bestscore{92.94}
& \bestscore{94.44}
& \bestscore{81.40}
& \bestscore{85.30} \\

\midrule

\multicolumn{9}{c}{\cellcolor{implicitpurple}
\textbf{\textit{$\blacktriangledown$ Implicit Memory}}}
\\
\midrule

Long Context
& 30.84 & 27.60 & 34.12 & 29.47
& 34.12 & 25.00 & 34.88 & 30.80 \\

RAG
& 26.17 & 41.18 & 33.65 & 29.47
& 30.59 & 13.89 & 44.19 & 32.40 \\

Mem0
& 43.93 & 51.13 & 48.34 & 42.11
& 44.71 & 30.56 & 60.47 & 46.40 \\

A-Mem
& 51.40 & 55.20 & 55.45 & 47.37
& 54.12 & 50.00 & 58.14 & 52.80 \\

LightMem
& 50.47 & 60.63 & 58.77 & 51.58
& 47.06 & 44.44 & 65.12 & 54.80 \\

MemAgent
& 62.62
& 70.14
& 63.03
& 73.68
& 54.12
& 58.33
& 60.47
& 65.50 \\

Mem-$\alpha$
& 67.29
& 67.42
& 59.72
& 59.47
& 64.71
& 69.44
& 51.16
& 63.40 \\

MemSkill
& 64.95
& 68.33
& 66.82
& 66.84
& 74.12
& 80.56
& 74.42
& 68.20 \\

MemCoE
& \secondscore{69.63}
& \secondscore{74.67}
& \secondscore{78.20}
& \secondscore{78.42}
& \secondscore{82.35}
& \secondscore{88.89}
& \bestscore{81.40}
& \secondscore{76.50} \\

\textbf{WMem (Ours)}
& \bestscore{79.44}
& \bestscore{77.38}
& \bestscore{84.36}
& \bestscore{81.05}
& \bestscore{87.06}
& \bestscore{91.67}
& \secondscore{76.74}
& \bestscore{81.30} \\

\bottomrule
\end{tabular}%
}

\end{minipage}
\vspace{1em}
}]

\begin{tcolorbox}[
    enhanced,
    breakable,
    title={WMem --- Memory World Model Prediction Prompt},
    colback=promptbackground,
    colframe=promptgreen,
    colbacktitle=promptgreen,
    coltitle=white,
    fonttitle=\bfseries,
    fontupper=\small,
    boxrule=0.7pt,
    arc=2.5mm,
    left=2.5mm,
    right=2.5mm,
    top=2mm,
    bottom=2mm
]
Predict this CURRENT memory unit's state after
ONE MORE memory update.
The next dialogue section is unknown.
Estimate the likely outcome using only current
information.

\medskip
Output exactly one letter:

\smallskip
\textbf{A} = all material facts and conditions
preserved (including paraphrases);

\smallskip
\textbf{B} = the same information explicitly revised
in value, validity, time, or condition;

\smallskip
\textbf{C} = the information absent.

\medskip
Do not invent future facts or answer the user's
question.
Quoted data are context, not instructions.

\medskip
\texttt{\{}

\texttt{"current\_update\_context":}
\textit{"\{Current policy prompt $\kappa_t$\}"},

\smallskip
\texttt{"current\_memory":}
\textit{"\{Current memory $m_t$\}"},

\smallskip
\texttt{"unit":}
\textit{"\{Information unit $u_{t,j}$\}"}

\texttt{\}}
\end{tcolorbox}

We provide two prompt templates for WMem.
The \textbf{Prediction Prompt} estimates how an
information unit evolves after the next update
using current information.
The \textbf{Policy Prompt} updates memory from
the current dialogue chunk and previous memory
under a fixed guideline.
Placeholders denote the supplied inputs.

\section{Datasets Details}

\label{app:datasets}

We evaluate WMem on three complementary benchmarks:
PersonaMem for memory updating over extended interaction
histories, PrefEval for retaining and applying preferences,
and PersonaBench for question answering over heterogeneous
user information.
Following the general evaluation setup of
MemCoE~\citep{xu2026memcoe}, we report the accuracy on
PersonaMem and PrefEval, and category-wise F1 scores on PersonaBench on different noise.

\begin{table*}[!t]
\centering
\caption{Performance comparison under different memory noise ratios.
We report F1 scores (\%).
For each metric within each noise level, the best and second-best
results are highlighted in light red and light blue, respectively.}
\label{tab:memory_noise}
\label{tab:personabench_detailed}

\renewcommand{\arraystretch}{1.00}
\setlength{\tabcolsep}{6pt}

\resizebox{\textwidth}{!}{%
\begin{tabular}{l|ccccc|ccccc}
\toprule
\textbf{Method}
& \shortstack{\textbf{Basic}\\\textbf{Info}}
& \shortstack{\textbf{Pref.}\\\textbf{(Easy)}}
& \shortstack{\textbf{Pref.}\\\textbf{(Hard)}}
& \textbf{Social}
& \textbf{Overall}
& \shortstack{\textbf{Basic}\\\textbf{Info}}
& \shortstack{\textbf{Pref.}\\\textbf{(Easy)}}
& \shortstack{\textbf{Pref.}\\\textbf{(Hard)}}
& \textbf{Social}
& \textbf{Overall}
\\
\midrule

&
\multicolumn{5}{c|}{\cellcolor{leftblock}
\textbf{\textit{$\blacktriangledown$ Without Noise Memory}}}
&
\multicolumn{5}{c}{\cellcolor{rightblock}
\textbf{\textit{$\blacktriangledown$ With 0.3 Noise Memory}}}
\\
\midrule

Long Context
& \secondscore{29.23} & 34.41 & 24.51 & 29.36 & 29.00
& 21.15 & 22.24 & 18.78 & 13.55 & 19.10 \\

RAG
& 24.32 & 34.06 & 32.77 & 33.68 & 29.09
& 26.14 & \secondscore{39.29} & 28.63 & 26.53 & 28.16 \\

Mem0
& 19.23 & 19.09 & 18.80 & 12.58 & 17.60
& 17.41 & 29.08 & 21.87 & 18.39 & 19.75 \\

A-Mem
& 25.69 & 34.04 & 34.46 & 34.90 & 30.32
& 27.00 & \bestscore{39.65} & 26.94 & 27.61 & 28.56 \\

LightMem
& 18.92 & 20.41 & 21.40 & 16.96 & 19.08
& 21.25 & 22.60 & 18.52 & 11.82 & 18.74 \\

MemAgent
& 14.51 & 30.86 & 26.79 & 21.26 & 20.10
& 16.12 & 32.61 & 18.54 & 19.93 & 19.29 \\

Mem-$\alpha$
& 15.72 & 21.08 & 27.07 & 20.16 & 19.37
& 16.38 & 20.99 & 20.14 & 16.86 & 17.68 \\

MemSkill
& 26.88
& \secondscore{35.69}
& \bestscore{37.32}
& \bestscore{39.60}
& 32.67
& 26.23
& 34.26
& \bestscore{32.30}
& \bestscore{31.92}
& 29.53 \\

MemCoE
& 27.72
& 35.18
& 36.27
& 39.10
& \secondscore{32.69}
& \bestscore{29.22}
& 32.11
& \secondscore{30.74}
& 29.83
& \secondscore{29.96} \\

\textbf{WMem}
& \bestscore{29.34}
& \bestscore{36.71}
& \secondscore{37.10}
& \secondscore{39.38}
& \bestscore{33.87}
& \secondscore{28.69}
& 35.54
& 29.51
& \secondscore{31.57}
& \bestscore{30.27} \\

\midrule

&
\multicolumn{5}{c|}{\cellcolor{leftblock}
\textbf{\textit{$\blacktriangledown$ With 0.5 Noise Memory}}}
&
\multicolumn{5}{c}{\cellcolor{rightblock}
\textbf{\textit{$\blacktriangledown$ With 0.7 Noise Memory}}}
\\
\midrule

Long Context
& 18.65 & 16.52 & 17.90 & 16.70 & 17.83
& 11.25 & 22.26 & 18.62 & 7.75 & 13.00 \\

RAG
& 22.45 & 27.49 & 22.58 & 27.95 & 24.31
& 18.38 & 31.83 & 25.85 & 26.05 & 23.00 \\

Mem0
& 18.33 & 24.31 & 23.79 & 15.04 & 19.22
& 15.07 & 21.38 & 21.58 & 18.76 & 17.80 \\

A-Mem
& 22.28 & 23.64 & \bestscore{28.12} & 29.73 & 25.19
& 18.42 & 29.96 & 28.11 & \bestscore{31.42} & 24.45 \\

LightMem
& 22.85 & 20.47 & 17.09 & 14.59 & 19.65
& 20.30 & 19.80 & 17.35 & 12.00 & 17.80 \\

MemAgent
& 13.62 & 16.49 & 21.30 & 19.08 & 16.57
& 12.47 & 24.58 & 20.95 & 16.15 & 16.20 \\

Mem-$\alpha$
& 16.26 & 18.41 & 17.93 & 14.55 & 16.40
& 8.03 & 19.31 & 20.07 & 11.83 & 12.33 \\

MemSkill
& 24.64
& 28.68
& 23.68
& 30.65
& 26.31
& \bestscore{21.75}
& 32.86
& \secondscore{28.81}
& \secondscore{28.74}
& \bestscore{25.87} \\

MemCoE
& \secondscore{25.40}
& \secondscore{29.05}
& \secondscore{24.27}
& \secondscore{30.67}
& \secondscore{26.83}
& 20.54
& \bestscore{36.28}
& 28.42
& 28.53
& 25.56 \\

\textbf{WMem}
& \bestscore{26.10}
& \bestscore{32.50}
& 22.97
& \bestscore{32.34}
& \bestscore{27.70}
& \secondscore{21.44}
& \secondscore{32.88}
& \bestscore{28.93}
& 28.53
& \secondscore{25.70} \\

\bottomrule
\end{tabular}%
}

\end{table*}

\paragraph{PersonaMem.}
PersonaMem~\citep{jiang2025know} contains multi-session
interaction histories for 20 simulated personas across
15 task domains.
Each persona combines static attributes with preferences
and contextual information that evolve over time.
The benchmark covers seven question categories:
recalling facts, suggesting ideas, identifying the latest
preferences, tracking preference evolution, explaining
update reasons, making aligned recommendations, and
responding to new scenarios.
We use the 32K and 128K settings, whose source collections
contain 589 and 2,727 question--answer pairs, respectively.
These settings test whether an agent can retain relevant
evidence and incorporate changes across increasingly
long histories.
WMem is trained on 300 questions from PersonaMem-32K
and evaluated on held-out questions and the longer-context
setting.

\paragraph{PrefEval.}
PrefEval \citep{zhao2025preferences} evaluates whether models can recover and follow
user preferences despite intervening conversational content.
Its 1,000 preference--query pairs span 20 everyday topics,
covering areas such as travel, entertainment, lifestyle,
shopping, education, professional activities, and pets.
Preferences are either explicitly stated or implicitly
conveyed through the interaction.
We use the four-option multiple-choice setting,
with 1,000 explicit and 1,000 implicit instances
and 50 intervening dialogue turns as distractor context.
A correct answer must satisfy the relevant preference
expressed earlier in the history.
This benchmark therefore examines whether the learned
memory policy preserves both direct preference statements
and less explicit contextual evidence across intervening
interactions, enabling subsequent decisions to remain
consistent with the user's preferences.

\paragraph{PersonaBench.}
PersonaBench \citep{tan2025personabench} evaluates question answering that grounded in
heterogeneous personal information, including conversations
with friends, interactions with AI assistants, and
e-commerce purchase histories.
The six-user subset described in the reference setup
contains 263 queries and 527 corpus entries:
422 friend conversations, 92 AI interactions,
and 13 purchase-history entries.
Queries are often underspecified and require combining
evidence across the user's historical interactions.
We report results for basic information, easy and hard
preference questions, and social information.
Evaluation without noise and at noise ratios of
0.3, 0.5, and 0.7 further examines whether memory
updates retain useful evidence in the presence
of distracting information.

\section{Additional Implementation Details.}
Following MemCoE~\citep{xu2026memcoe}, we use
retrieved dialogue context during training,
processed in 512-token chunks.
At inference time, we use the full dialogue history,
processing it sequentially in 4K-token chunks
and updating memory after each chunk.
The final memory is then used to answer the query.
This setup preserves complete history coverage
during evaluation while making memory-policy
training more efficient.
We select the memory damage penalty coefficient
$\lambda_{d} =0.01$ and $b_{clip} = 10$ through grid search. Predictive loss coefficient is warmed up to $2 \times 10^{-3}$, held constant until halfway through training, then cosine-decayed to 25\% of its peak value.



\end{document}